# A Survey on Moral Foundation Theory and Pre-Trained Language Models: Current Advances and Challenges

Lorenzo Zangari[a], Candida M. Greco[a], Davide Picca[b], Andrea Tagarelli[a]

[a] *University of Calabria, Rende (CS), 87036, Italy*
[b] *University of Lausanne, Lausanne, 1015, Switzerland*

## Abstract

Moral values have deep roots in early civilizations, codified within norms and laws that regulated societal order and the common good. They play a crucial role in understanding the psychological basis of human behavior and cultural orientation. The Moral Foundation Theory (MFT) is a well-established framework that identifies the core moral foundations underlying the manner in which different cultures shape individual and social lives. Recent advancements in natural language processing, particularly Pre-trained Language Models (PLMs), have enabled the extraction and analysis of moral dimensions from textual data. This survey presents a comprehensive review of MFT-informed PLMs, providing an analysis of moral tendencies in PLMs and their application in the context of the MFT. We also review relevant datasets and lexicons and discuss trends, limitations, and future directions. By providing a structured overview of the intersection between PLMs and MFT, this work bridges moral psychology insights within the realm of PLMs, paving the way for further research and development in creating morally aware AI systems.

*Keywords:* Pre-Trained Language Models, Moral Foundation Theory, Moral Values Survey, Natural Language Processing

*Email addresses:* `lorenzo.zangari@dimes.unical.it` (Lorenzo Zangari), `candida.greco@dimes.unical.it` (Candida M. Greco), `davide.picca@unil.ch` (Davide Picca), `andrea.tagarelli@unical.it` (Andrea Tagarelli)

## 1. Introduction

Moral values have roots in the earliest civilizations, which had already codified ethical principles within norms and laws that regulated daily life, attesting to a deep reflection on the order of society and the common good. Socrates, Plato, and Aristotle explored the nature of virtues and ethics, linking them with the exploration of individual and collective well-being, which can help people define who they are and what they aspire to be, and are foundational elements for moral values. Indeed, moral values guide how to behave, not only personally but also as the pillars of constructing and maintaining social order. They instill a sense of responsibility and encourage behaviors that would support the well-being of the community (Schwartz, 1992) or increase the division between people within society (Brady et al., 2020). This makes morality a distinctive element of individual character, which influences decisions and human behavior that can lead to a sense of integrity and self-esteem (Rana and Solaiman, 2022), and to implication of what is a morally good or bad life (Hofmann et al., 2014; Ellemers, 2018; Kádár et al., 2019).

In recent years, the integration of moral foundations into computational models has been an area of active research. The advances in Natural Language Processing (NLP) have revolutionized the way we extract information and knowledge from textual data. Particularly, *Pre-trained Language Models (PLMs)*, powered by Transformer backbones (Vaswani et al., 2017), have become a groundbreaking technology in addressing a wide range of tasks (Lin et al., 2022). Transformers' self-attention mechanism allows for capturing global complex dependencies within input data, enabling exceptional performance in NLP and beyond. In this respect, one of the emerging trends is to integrate PLMs with moral information for inferring moral foundations from the input data and analyzing the moral dimensions embedded in PLMs. These approaches are beneficial for applications in social sciences, policy-making, and even in enhancing AI ethics. For example, PLMs can be employed to execute moral judgment tasks like moderating content on social platforms (Franco et al., 2023); or to analyze political speeches and campaigns, as well as business strategies, identifying potential manipulative tactics (Liu et al., 2022; Luceri et al., 2024). Hence the need to study AI systems for morality becomes paramount. Furthermore, recent studies have shown that PLMs can be trained to recognize and reflect moral judgments, or can be inspected to reveal certain moral tendencies present in the pre-training datasets (Guo et al., 2023a; Preniqi et al., 2024; Zhang et al., 2023). Therefore, the study of methodologies integrating PLMs and moral foundations can provide a kind of transparency that becomes the basis for user's confidence in AI systems. Another significant stride in this direction has been the creation of large-scale datasets enriched with moral dimensions (Hoover et al., 2020; Trager et al., 2022), and specialized lexicons (Araque et al., 2020) enabling a nuanced analysis of moral contents in textual data. These data sources are essential for training and/or fine-tuning PLMs for moral content analysis. However, the reliability of such data sources is called into question. Moral concerns are differentially observable in language, and context-dependent moral judgments often elude simple lexicon-based detection methods (Kennedy et al., 2021).

In this work, we provide a systematic overview of the recent advances in the integration of moral foundations within PLMs, from the perspective of the *Moral Foundation Theory (MFT)* (Haidt and Joseph, 2004). The MFT posits that human moral reasoning is based on innate, modular foundations, and identifies five core dyadic moral foundations: Care/Harm, Fairness/Cheating, Loyalty/Betrayal, Authority/Subversion, and Sanctity/Degradation. MFT is widely regarded as a robust and versatile theory since it encapsulates a broad spectrum of moral values that are pertinent across different cultures and societies. This makes MFT particularly suitable for applications that need to operate in diverse and multicultural environments. Furthermore, MFT provides a structured approach to decomposing and analyzing moral values, which is crucial for evaluating the behavior of PLMs in moral scenarios.

*Objectives and scope of the survey.* In this survey, we aim to analyze the current approaches to combining the MFT and PLMs to systematize and provide valuable insights into how PLMs perceive and generate relevant content within a moral framework, like the MFT. The increasing use of PLMs in diverse applications underscore the critical nature of this aspect. We summarize our main contributions as follows:

- **New taxonomy**: We provide a comprehensive overview of the MFT and the PLMs involved in the context of the MFT. Additionally, we formalize the essential strategies and tasks adopted within the domain of MFT-informed PLMs. Based on our theoretical framework, we organize the existing works in the literature into two main categories: (i) *Moral-driven PLMs*, which focus on training PLMs on MFT-based data; and (ii) *Moral-targeted PLMs*, which analyze the moral foundations embedded in PLM responses at the inference phase.

- **Moral Surveys, Lexicons, and Datasets**: We review and categorize the most commonly used datasets, surveys and lexicons for training and assessing PLMs on MFT-related tasks.
- **Findings, Challenges, and Future Directions**: We discuss key findings from the reviewed works, highlight ongoing challenges in the field, and suggest future research directions. We emphasize the importance of constructing reliable and inclusive datasets, along with the pivotal role of PLMs in the context of MFT. This includes considering aspects such as the model size and the strategies employed for training or fine-tuning the PLMs.

By providing a structured overview of the intersection between the fields of PLMs and MFT, this survey aims to bridge the gap between moral psychology and advanced computational linguistics, laying the groundwork for future research and development in the creation of morally aware AI systems.

Although there exist similar works as discussed in Section 2, to the best of our knowledge, this is the first intellectual effort to introduce and provide a comprehensive review of the combination of the MFT and PLMs. The remainder of this work is organized as follows. Section 2 summarizes existing surveys on PLMs which focus on moral, ethics and cultural values aspects. Section 3 provides a theoretical background of the two pillars of this survey: MFT and PLMs. In section 4 we formalize the tasks arising from the integration of MFT and PLMs. Section 5 gives an overview of widely adopted moral data sources. In section 6 we present our taxonomy and describe existing works. In section 7 we discuss the main trends, challenges, and possible future directions in the field.

## 2. Related works

Recent advancements in the field of moral values within PLMs include the design of algorithms capable of detecting morally relevant contents and the implementation of procedures or evaluation metrics that reflect the moral perceptions of PLMs. However, the existing literature lacks comprehensive reviews that explore the intersection of MFT and PLMs within a computer science perspective. Existing surveys mainly focus on social science, evaluating the integration of PLMs within human society from a broader outlook. For instance, Adilazuarda et al. (2024) investigate cultural values representation and inclusion in PLMs, highlighting the absence of a consistent definition of culture and employing different proxies to represent cultures. They categorized two broad approaches to studying cultural values inside PLMs: observing the response of the PLMs to various inputs and observing the internal state of PLMs. Wang et al. (2023) review the existing alignment tools for PLMs, with emphasis on training approaches and model assessment strategies, pointing out the limits of PLMs in humans, such as misreading human commands and producing biased or erroneous material. Guo et al. (2023b) offer a review of the evaluation methods of PLMs, including aspects related to morality, such as the typically used datasets related to the MFT. Similarly, Chang et al. (2024) provide an overview of the evaluation methods for PLMs in terms of robustness, ethics, biases, and trustworthiness. Gallegos et al. (2023) summarize bias evaluation and mitigation techniques for PLMs, expanding the notion of fairness and harm to improve the fairness capabilities of PLMs.

*Differences with existing surveys.* We systematically review works that are framed within the context of PLMs and the MFT. Compared to other works based on the combination of PLMs and human psychological aspects, such as Adilazuarda et al. (2024), we do not focus solely on reviewing methods that observe the responses or the status of the PLMs, but we also consider fine-tuned or further pre-trained PLMs for detecting moral values. Similarly to Wang et al. (2023), we aim to review the evaluation and training strategies, but for the specific area of MFT and PLMs. Guo et al. (2023b); Chang et al. (2024) and Gallegos et al. (2023) investigate moral values only from an evaluation perspective and within the much broader scope of ethics, safety, and AI alignment. To the best of our knowledge, there is a lack of literature on surveys that review works based on MFT-informed PLMs. We believe that our survey can bridge the gap by providing a focused examination of the intersection between MFT and PLMs, offering insights into how these theoretical frameworks can be applied to enhance and comprehend the behaviour of PLM on morally-relevant data. By integrating insights from a computer science perspective, we aim to provide a comprehensive overview that highlights the potential and limitations of current methodologies and suggests directions for future research.

## 3. Theoretical Background

In this section, we describe the two fundamental pillars of this survey: the Moral Foundation Theory and Pre-trained Language Models. In section 3.1 we delve into the principles and applications of the MFT, while in section 3.2 we provide an overview of the PLMs involved in the works explored in this survey, as well as a brief description of the strategies they employed for tailoring PLMs to downstream applications.

### *3.1. Moral Foundation Theory*

Moral Foundations Theory was formulated by Haidt and Joseph (2004) to investigate the underlying reasons behind common moral themes across different cultures. It claims that multiple psychological systems form the basis of our intuitive ethics, which are then shaped by cultures' virtues, narratives, and institutions. MFT does not judge the moral value of these systems, rather it provides a descriptive analysis of human morality based on 4 pillars: (i) *Nativism*, which involves the hypothesis that certain moral foundations are innate; (ii) *Cultural learning*, which is the process that enables individuals to acquire moral values from their culture; (iii) *Intuitionism*, which corresponds to the notion that moral judgments are often made quickly and automatically; and (iv) *Pluralism*, which recognizes that there are multiple, sometimes conflicting, moral principles that people adhere to. If any of these pillars were to be proven false, the MFT would no longer be valid (Graham et al., 2013).

MFT's initial framework established five fundamental principled dichotomies within this evolutionary context, or *moral foundation dimensions*. Each of these dimensions represents a critical aspect of social cooperation and conflict resolution that would have been essential for our ancestors' survival and flourishing:

- **Care (vs. Harm)**: making pressure to care for others and repel hurting others as the basis of this foundation. It prompts us to care and be kind to people who are suffering in vulnerable positions, such as those who take refuge, animals, or even those who perform in movies.
- **Fairness (vs. Cheating)**: centering on mutual altruism, it correlates with the rule of fairness and treatment of social justice, lauding virtues like honesty and integrity.

| Foundation | Care/harm | Fairness/cheating | Loyalty/betrayal | Authority/subversion | Sanctity/degradation |
|---|---|---|---|---|---|
| Adaptive challenge | Protect and care for children | Reap benefits of two-way partnerships | Form cohesive coalitions | Forge beneficial relationships within hierarchies | Avoid communicable diseases |
| Original triggers | Suffering, distress, or neediness expressed by one's child | Cheating, cooperation, deception | Threat or challenge to group | Signs of high and low rank | Waste products, diseased people |
| Current triggers | Baby seals, cute cartoon characters | Marital fidelity, broken vending machines | Sports teams, nations | Bosses, respected professionals | Immigration, deviant sexuality |
| Characteristic emotions | Compassion for victim; anger at perpetrator | Anger, gratitude, guilt | Group pride, rage at traitors | Respect, fear | Disgust |
| Relevant virtues | Caring, kindness | Fairness, justice, trustworthiness | Loyalty, patriotism, self-sacrifice | Obedience, deference | Temperance, chastity, piety, cleanliness |

Figure 1: The original five moral foundations Graham et al. (2013). The Foundation column refers to the basic building blocks or core principles of human morality.

- **Loyalty (vs. Betrayal)**: the concept brings to light the key aspect of group or tribe loyalty which always resulted in survival advantages when organizing strong relational bonds.
- **Authority (vs. Subversion)**: the organization's underlying principle is that of conservatism, emphasizing the importance of respect and the legitimacy of authorities and traditions. Such values as obedience and deference are evaluated differently around the world.
- **Sanctity (vs. Degradation)**: being a narrower definition, this foundation is based on the concern with purity and pollution and it is related to the 'behavioral immune system,' which serves as a guide to our reactions to potential sources of infection or moral corruption.

Care, Fairness, Loyalty, Authority and Sanctity are categorized as *Virtues*, while their counterparts are referred to as *Vices*. Figure 1 summarizes the key aspects behind each moral dimension of the MFT. Each foundation is connected with the following 5 basic principles of human morality (Graham et al., 2013): (i) Adaptive challenge: The basic survival need or challenge that each foundation addresses. (ii) Original triggers: The initial situations or stimuli that originally activated each moral foundation. (iii) Current triggers: Modern scenarios or stimuli that currently activate each moral foundation. (iv)

Characteristic emotions: The typical emotions associated with each moral foundation. (v) Relevant virtues: The virtues or positive character traits linked to each moral foundation.

Additional foundations have been proposed to reflect ongoing research and societal changes (Haidt, 2012), such as **Liberty/Oppression**, which emphasizes the feelings against domination and the drive for autonomy, often clashing with authority-related intuitions.

The pluralistic approach of MFT has garnered significant empirical support, with studies demonstrating both its cross-cultural validity and predictive power.

A landmark study by Graham et al. (2011) provided robust evidence for the cross-cultural applicability of MFT. Their comprehensive research, employed the Moral Foundations Questionnaire (MFQ) to assess the relevance and endorsement of each moral foundation. The study's findings strongly supported the universality claim of MFT, with the five-factor model of moral foundations consistently emerging across diverse cultures. Notably, while the overall structure of moral foundations remained stable, the degree of endorsement varied across cultures. Eastern cultures, for instance, tended to show a more balanced emphasis across all five foundations, whereas Western cultures often prioritized the individualizing foundations (Care and Fairness) over the binding foundations (Loyalty, Authority, and Sanctity).

Building on this cross-cultural validation, Nilsson and Erlandsson (2015) further bolstered the empirical support for MFT by demonstrating its superior predictive power for moral judgments compared to competing unidimensional models. Their study, conducted in Sweden, employed structural equation modeling to compare MFT with alternative models such as the Moral Identity Model and the Model of Moral Motives. Their study employed confirmatory factor analysis to compare the five-factor model of MFT with alternative two- and three-factor models. The results confirmed that the five-factor structure of MFT provided a better fit than the alternatives, even in the highly secular Swedish society. However, the authors noted that the fit was not optimal, suggesting potential areas for refinement in the measurement of moral foundations across different cultures.

Furthermore, the study explored how the relationship between moral foundations and political ideology was mediated by factors such as preference for equality, resistance to change, and system justification. This analysis provided new insights into the mechanisms linking moral intuitions to political attitudes in the Swedish context. While supporting the overall structure of MFT, the authors also pointed out some psychometric challenges with the Moral Foundations Questionnaire, particularly in non-English versions. This suggests that further refinement of the measure may be necessary to fully capture the nuances of moral foundations across different languages and cultures.

While these studies collectively provide robust empirical support for MFT, demonstrating both its cross-cultural validity and predictive power, they also reveal important nuances in how moral foundations are expressed across different cultures. This dual nature of MFT – its apparent universality and its cultural variability – emphasizes the importance of considering both universal and culture-specific aspects of morality in its application.

Overall, MFT provides a robust framework for exploring how innate moral intuitions, shaped by evolutionary and cultural factors, guide human behavior and social interactions. It has been extensively applied within social psychology to explain a variety of empirical findings and has significant implications beyond this field. We emphasize that, in this work, we discuss morality within the framework of MFT, thereby selecting works that how these principles can be applied when integrated with PLMs.

### *3.2. Pre-trained Language Models*

Pre-trained language models have profoundly transformed the field of NLP, pushing society into an era of sophisticated language comprehension and generation. Trained on extensive corpora, these models discern intricate patterns and structures within language, enabling them to excel in tasks such as translation, summarization, and question-answering. At the core of these capabilities lies the self-attention mechanism of Transformer architecture, which serves as the fundamental component driving the performance and versatility of various PLMs.

In this Section, we describe the specific PLMs employed in the works discussed in section 6. We also provide a brief summary of the main strategies for adapting PLMs to downstream tasks. We emphasize that the discussed strategies are not exhaustive but are specifically those employed by the studies considered in this survey to address MFT-related tasks. Figure 2 shows a diagram of the potential PLMs architectures and strategies that can be employed by these architectures in the field of MFT and PLMs. In the following, we provide an overview of the PLMs involved in the works described in section 6 organized by architecture type (e.g., Encoder, Encoder-Decoder).

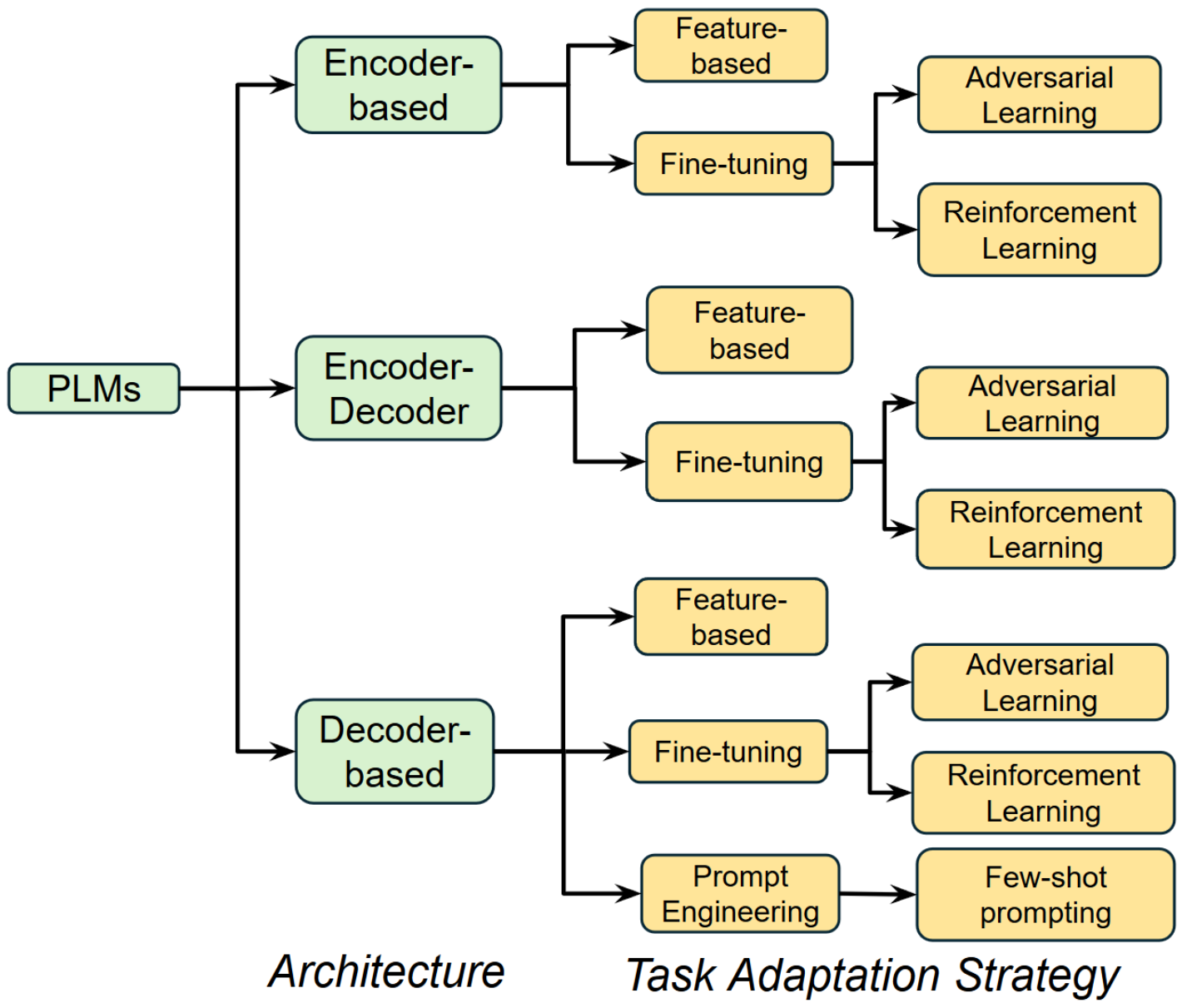

Figure 2: Diagram of PLMs architectures with relative task adaptation strategies.

*3.2.1. Encoder-based models: BERT and BERT-based*

BERT-based models are a family of PLMs that leverage the Bidirectional Encoder Representations from Transformers architecture (Devlin et al., 2019). These models are encoder-based, i.e., they primarily focus on encoding the input text into dense vector representations. This approach allows the models to capture global and contextual relationships within the text, making them highly effective for understanding and processing language.

*BERT* (Devlin et al., 2019) has set a new standard in NLP by introducing the bidirectionality in the pre-training phase. Unlike previous models that read text sequentially, BERT reads the entire sequence of words at once by considering simultaneously the left and right context, hence allowing it to deeply understand the entire input. As a result, BERT has significantly improved performance across a variety of NLP tasks, including named entity recognition, question answering, and language inference. BERT is released in two different sizes: BERT-*base* and BERT-*large*. BERT-*base* comprises 12 stacked encoder layers, each with 12 attention heads and a hidden size of 768, totaling 110M parameters. In contrast, BERT-*large* includes 24 stacked encoder layers with 16 attention heads and a hidden size of 1024, resulting in a total of 340M parameters. Over the years, several instances of the BERT models have been released. These versions primarily differ in model size (*base* or *large*), as well as in training objectives and case folding (*cased* or *uncased*).

*S-BERT* (Reimers and Gurevych, 2019) fine-tunes BERT to generate high-quality sentence embeddings efficiently, which are particularly useful for tasks like semantic search, sentence similarity, and clustering. S-BERT significantly reduces the computational effort required by BERT to find the most similar pair in a collection while maintaining the same level of accuracy. Currently, more than 5000 models inspired by S-BERT (better known as Sentence-Transformer) are available,[1] with any PLM serving as a backbone model.

*SqueezeBERT* (Iandola et al., 2020) is a variant of BERT designed to be more efficient by reducing latency and computational requirements while maintaining comparable performance levels, particularly for use on mobile devices and other environments with limited computational resources. It contains the same number of layers, attentions heads and hidden layers as BERT-*base*, but with half the total parameters (51.1M parameters).

#### *3.2.2. Encoder-Decoder based models: T5*

Unlike BERT, which is primarily an encoder model, T5 (Raffel et al., 2020) employs an encoder-decoder architecture that allows it to handle a wide variety of text-related tasks within a unified framework. The key innovation of T5 is its text-to-text approach, where all tasks are framed as text generation problems. Thus, the inputs and outputs for every task are represented as text strings, simplifying the model's application across different NLP tasks. T5 is available in five versions: T5-*small* (6 layers, 8 attention heads, 512 hidden size, 60M parameters), T5-*base* (12 layers, 12 attention heads, 768 hidden size, 220M parameters), T5-*large* (24 layers, 16 attention heads, 1024 hidden size, 770M parameters), T5-3*B* (24 layers, 32 attention heads, 1024 hidden size, 3B parameters), T5-11*B* (24 layers, 128 attention heads, 1024 hidden size, 11B parameters).

[1] `https://sbert.net/index.html`

*3.2.3. Decoder-based models: Large Language Models*

Other models such as GPT-3, GPT-3.5 (Brown et al., 2020), ChatGPT (OpenAI, 2023), and open alternatives like LLaMA-2 (Touvron et al., 2023) have further pushed the boundaries of what PLMs can achieve. Unlike BERT and T5, these models are decoder-based. Therefore, they can predict the next word in a sequence given the previous words, which makes them suitable for text generation tasks. These models are known as *Large Language Models (LLMs)* due to their extensive scale in terms of parameters (billions), training data, and computational resources, which enable them to perform a wide range of complex language tasks with high proficiency.

*GPT-3 and GPT-3.5* (Brown et al., 2020) are among the most advanced in this category, capable of producing human-like text. It is trained on text-based and code-based data and then aligned using Supervised Fine-Tuning (SFT), through which the model is trained to generate responses w.r.t. the format expected by users, and Reinforcement Learning from Human Feedback (RLHF) (Kalyan, 2024), to align the model behavior with human instructions and preferences. GPT-3 is available in different sizes, such as: *ada* (350M parameters), *babbage* (1.3B parameters), *curie* (6.7B parameters), *davinci* (175B parameters). GPT-3.5 is an improved version of GPT-3 with less parameters than *davinci* (20B parameters). ChatGPT (OpenAI, 2023) has GPT-3.5 (and GPT-4) as a core model, but is specifically optimized for interactive conversations and dialogues. It is designed to maintain the context of a conversation over multiple turns, making it suitable for interactive applications such as chatbots and virtual assistants.

*LLaMA-2* (Touvron et al., 2023) is an open-source model designed to enhance efficiency and performance in various NLP tasks, emphasizing open accessibility and robustness. Similarly to ChatGPT, it is optimized for chat-based interactions through SFT and RHLF. It demonstrates proficiency in both conversational and completion tasks and competitiveness compared with GPT-3.5. LLaMA-2 is available in multiple versions, including LLaMA-2 7B (7 billion parameters), LLaMA-2 13B (13 billion parameters), and LLaMA-2 70B (70 billion parameters).

*OPT* (Liu et al., 2021) is an open generative model designed to provide an accessible alternative to proprietary models like GPT-3.5, replicating its performance and sizes. OPT is available in different sizes, from 125M to 175B parameters.

*OpenAssistant 12B* (Köpf et al., 2023) is another open LLM containing 13 billion parameters. It is fine-tuned with reinforcement learning on human preference and supervised learning on question-answering and dialogue demonstrations. The fine-tuning is carried out on the OpenAssistant Conversations data, which is an assistant-style conversation corpus constructed with human assistance and annotations.

*Claude 3 Opus* (Anthropic, 2024) is an advanced language model recently designed for enhanced conversational AI capabilities, expert-level reasoning, complex content creation, and multilingual capabilities. The model is designed for reliability and reduced biases, ensuring safer and more responsible AI interactions.

*3.2.4. Strategies for tailoring PLMs to downstream applications*

In this section, we will analyze and explore the strategies utilized by the approaches discussed in Section 6 to tailor PLMs to MFT-related tasks.

*Fine-tuning.* The fine-tuning process involves additional training on a smaller, task-specific dataset, allowing the model to learn the nuances and requirements of that particular task while leveraging the broad knowledge it gained during pre-training. Specifically, it consists of training some or all layers of the PLM, then adding a task-specific layer (e.g., feed-forward output layers for classification), and training the model end-to-end. Early methods such as BERT need to be fine-tuned to perform a specific task, e.g., text classification. Fine-tuning is useful for various reasons, such as exposing the model to new or proprietary data (not included during pre-training, so there is a need to include in-domain knowledge) or to align the model's responses with human expectations when providing instructions (through for example fine-tuning with human feedback (OpenAI, 2023)).

*Feature-based approach.* The simplest way to use PLMs is to freeze their weights and use their output as context-sensitive word embeddings for a new architecture, which is specifically trained for the downstream task (Min et al., 2024) . This approach is similar to feature extraction in classic statistical NLP. Frozen contextual embeddings are employed in scenarios in which fine-tuning PLMs is impractical, e.g. when there is insufficient labeled data or for addressing computationally expensive NLP tasks.

*Prompt Engineering.* Prompting involves adding natural language text, typically in the form of short phrases, to provide guidance to pre-trained models in performing specific tasks (Min et al., 2024). This method has several advantages. First, prompting can avoid updating the PLM's parameters, reducing computational needs compared to fine-tuning. Second, prompts align the new task with the pre-training objective, enhancing the use of pre-trained knowledge and enabling few-shot learning with small training datasets. Finally, prompts allow unsupervised probing of PLMs to assess their knowledge on specific tasks. The most common type of prompt-based learning is few-shot learning, in which task descriptions and examples (0 or $k$ examples, respectively 0-shot or $k$-shot learning) are used to guide a PLM in performing the task.

*Adversarial Learning.* Adversarial learning is a technique that involves training deep learning models while exposing them to adversarial attacks, which are intentionally manipulated inputs designed to induce inaccuracies in the model's predictions (Goodfellow et al., 2015). By training the model on these crafted examples, it learns to be more resilient to adversarial attacks and less sensitive to small perturbations in the input. Typically, a min-max optimization framework trains the model to minimize its loss on adversarial examples and maximize its loss on the same examples to simulate adversarial conditions. It has been shown that this technique is highly beneficial for various NLP tasks (Yoo and Qi, 2021), such as text classification (Miyato et al., 2017). Also, adding adversarial examples during fine-tuning helps models be more general and reliable, even when dealing with complex, noisy, or adversarially manipulated data (You et al., 2023).

*Reinforcement Learning.* Reinforcement Learning (RL) is an advanced machine learning paradigm that teaches models to make sequences of decisions by rewarding desired behaviors and penalizing undesired ones. RL is particularly effective for aligning PLMs with human preferences. For instance, PLMs can be fine-tuned based on rewards derived from human evaluations of their outputs. This approach has been successfully employed to reduce harmful and biased responses, ensuring that PLMs generate outputs that are not only accurate but also align with societal norms and values (Stiennon et al., 2020). By leveraging human feedback, RL can adjust the behavior of PLMs more precisely than traditional supervised learning methods.

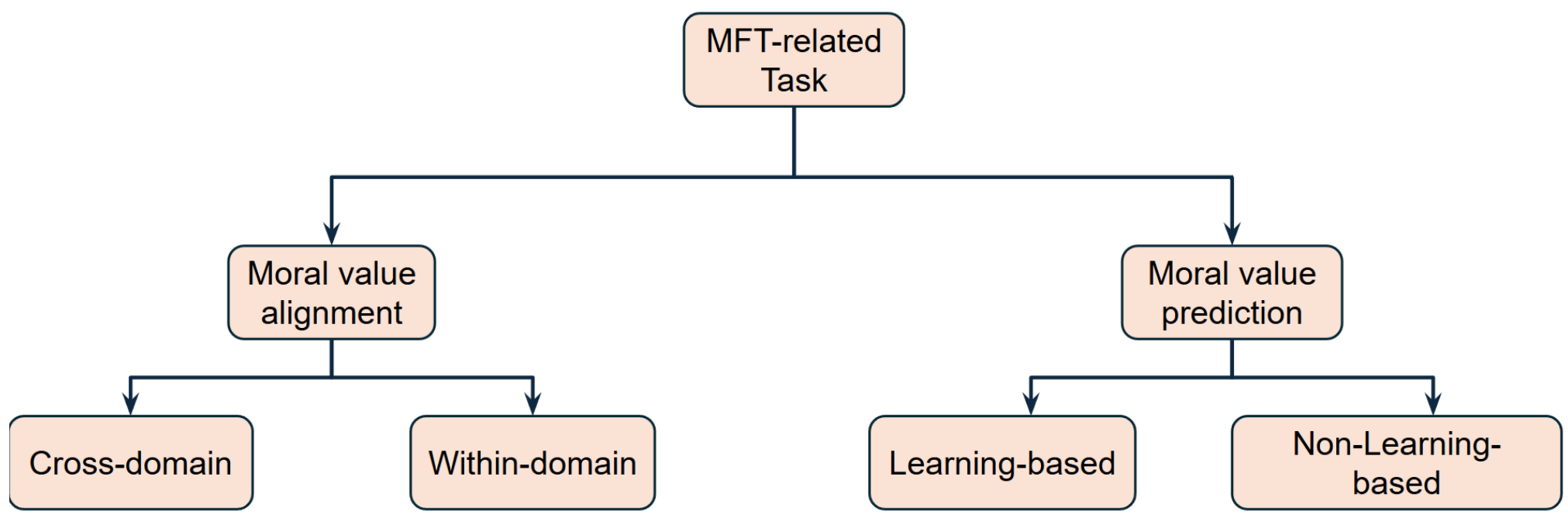

Figure 3: Categorization of the main MFT-related tasks addressed by existing literature.

## 4. MFT-related tasks

In this section, we discuss and formalize the tasks related to the integration of PLMs and MFT. Our objective is to present comprehensive and formal definitions of the primary tasks addressed by the investigated works, serving as a valuable foundation for future research endeavors. We categorize the identified tasks as illustrated in Figure 3. We emphasize that, to the best of our knowledge, this is the first survey to provide a taxonomy of the main MFT tasks faced by current research on PLMs.

We define $T_D$ as the MFT and $D$ as the set of *moral foundation values* associated to $T_D$. Without loss of generalizability, we assume that $D$ does not strictly represent the original MFT foundations with virtues and vices, bur rather moral values in which virtues and vices can be treated separately, i.e., $D = \{Care, Harm, Fairness, \dots\}$. Following this approach, $D$ could be easily integrated with other moral values, i.e. $\{\text{non-moral}\} \cup D$ (Hoover et al., 2020), which can be useful for learning more expressive models.

### *4.1. Moral values prediction*

Moral values prediction involves learning a function $M_\Theta$ that infer the moral values $D$ from the input data. Formally, given the input space $X$, the objective of the moral values prediction task is to learn $M_\Theta : X \mapsto Y$, where $Y \subseteq \mathcal{P}(D)$, and $\mathcal{P}(D)$ indicates the powerset of $D$. Note that $M_\Theta$ could be a PLM that is freezed, fine-tuned or further pre-trained (cf. section 3.2.4) for performing the final downstream task.

In order to train the PLM for the moral values prediction task, we assume the availability of a labeled dataset $L = (x_i, d_i)_{i=1}^{n}$, where $x_i \in X$ is an input instance, $d_i \subseteq D$ is the corresponding label(s), and $n$ is the size of the dataset. The goal is to minimize the loss function $\mathcal{L}(M_\Theta(x_i), d_i)$ over the labeled dataset, where the choice of the loss function depends on the particular setting of the problem (e.g., single-label or multi-label classification).

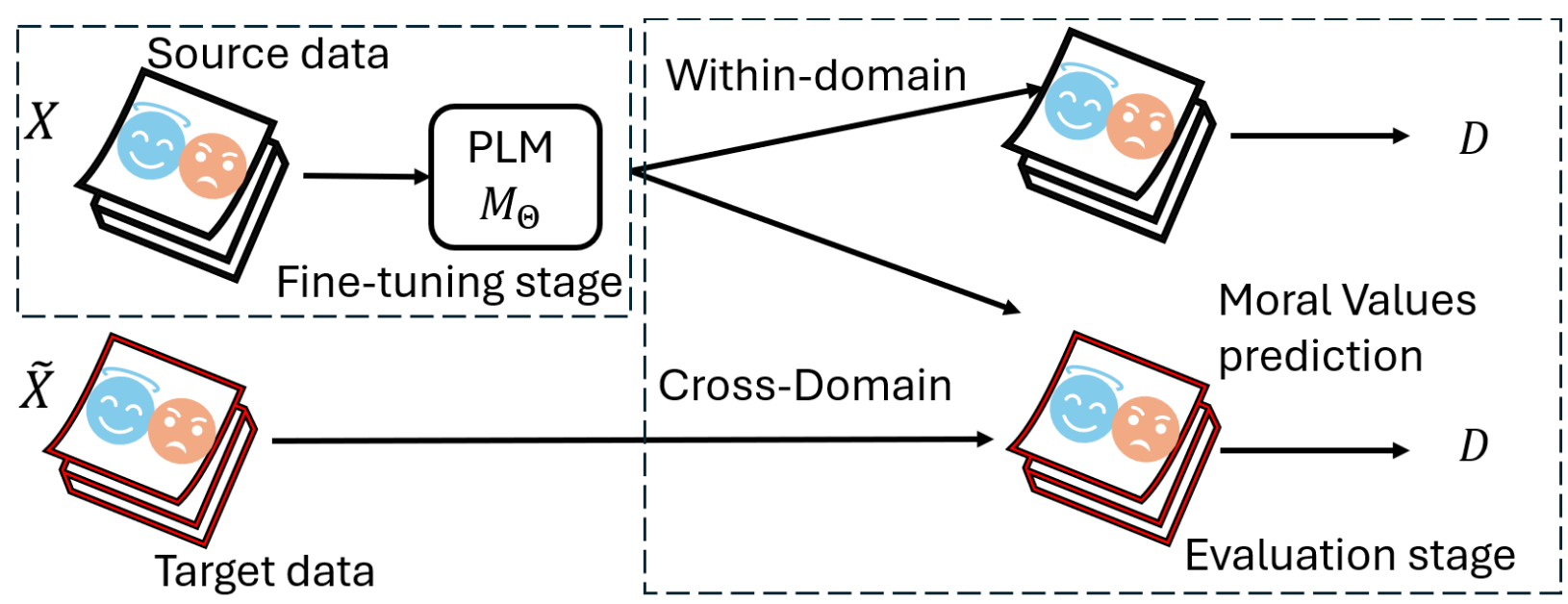

Figure 4: Within-domain and Cross-domain moral values prediction task. In the within-domain scenario, the model is trained and tested on data originating from the same distribution (source). In the Cross-domain scenario, the model is trained and tested on data originating from different distributions (source and target).

*Single-label classification*: In this case, each input instance is associated with exactly one moral value. The output space is defined as $Y = D$, and the function $M_\Theta$ maps each input instance to a single moral dimension, i.e., $M_\Theta(x_i)$ is a binary vector with 1 only in the entry corresponding to the moral value denoted by $d_i$.

*Multi-label classification*: In this setting, each input instance can be associated with multiple moral dimensions. The output space is defined as $Y = \{0,1\}^{|D|}$, where each element of the output vector represents the presence or absence of a particular moral dimension. The function $M_\Theta$ maps each input instance to a binary vector where $M_\Theta(x_i)_j = 1$ indicates that the $j$-th moral value is associated with the input instance $x_i$.

*Cross-domain and within-domain prediction.* Methods that address the moral values prediction task can be framed into within-domain and cross-domain scenarios. Note that in the current literature these scenarios can be referred to as out-of-domain and in-domain (Preniqi et al., 2024; Guo et al., 2023a; Liscio et al., 2022). Since all the works analyzed in this survey intend the same concept fro cross- (within-) and out-of- (in-) domain, hereinafter we will use the terms cross-domain and within-domain to create various setup that enrich that previously defined tasks, such as *cross-domain moral values* *prediction* and *within-domain moral values prediction.*

Within-domain methods focus on optimizing performance within the same domain of the training data. Specifically, train and test data are assumed to be drawn from the same distribution. Cross-domain approaches focus on transferring knowledge from the source domain(s) to the target domain. They involve adapting a model trained on one or multiple domains to perform well in a different, often less-resourced, domain. in which instances ($\tilde{X}$ are drawn from a different distribution w.r.t. the source samples $X$ This adaptation is crucial when source training data are scarce or expensive to obtain, making it challenging to train a model solely on the target domain data. By leveraging data from multiple domains during training, cross-domain approaches enhance the model's robustness to domain shifts and improve its generalization capabilities. Figure 4 schematize how a moral values prediction task is typically addressed.

### *4.2. Moral values alignment*

In the moral value alignment problem the goal involves analyzing whether the response of PLMs is aligned with specific moral values $D$. In this context, alignment refers to the evaluation of whether the response of the PLM is in line with the moral values expressed in the input text $X$ and denoted by $D$. Hereinafter, we will assume that the alignment function $f$ needs to be minimized, indicating that a lower value corresponds to a higher level of moral alignment for the PLM. The operational framework for addressing a moral value alignment task is shown in Figure 5. In our formulation, the problem can be approached as follows.

*Learning-based Moral Value Alignment*: We provide a similar setting to the one proposed by Dognin et al. (2024). Let us denote with $S$ the current conversation history of the PLM, $P$ the set of possible prompts given to the model, $R$ the set of responses that can be generated, and $f : R \times D \mapsto \mathbb{R}^k$ an alignment function. The objective is to learn a function $M_\Theta : S \times P \mapsto R$ that provides responses $r_i \in R$ for a given prompt in a given state such that $f(r_i, d_i)$ is minimized over the dataset $L = (r_i, d_i)_{i=1}^n$, with $d_i \in D$.

For example, if a PLM is prompted with scenarios involving a moral dilemma, the aligned responses would be those that consider the moral dimensions relevant to the scenario. Suppose a scenario that involve choosing between honesty and loyalty, such as deciding whether to tell the truth about a friend's mistake at work or to protect your friend by keeping it a secret. The responses chosen by the PLM should reflect an understanding of the moral weight of honesty (from the Fairness dimension) versus loyalty (from the Loyalty dimension) and make a decision that appropriately consider both values.

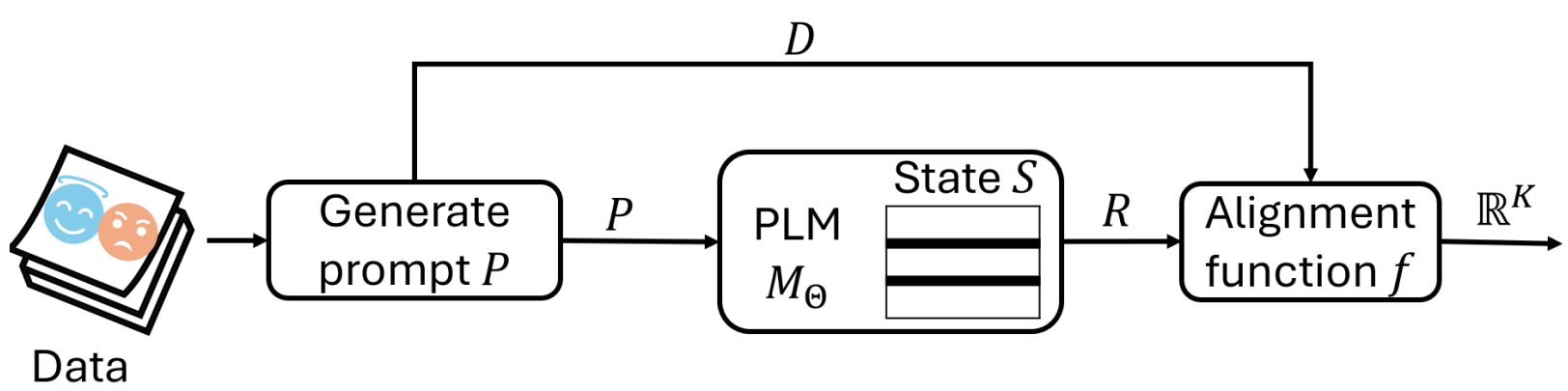

Figure 5: Moral values alignment problem. Given data with moral contents, a prompt expressing specific moral values $D$ is generated and input to a PLM, which has a state $S$ (e.g., conversation history). The moral values $D$ and the PLM's response are then fed into the alignment function $f$ to evaluate how well the PLM aligns with $D$. If the PLM is only used for inference, we are dealing with a Non-learning-based moral value alignment problem. If the PLM includes components that are trained or fine-tuned to align with $f$, we are dealing with the Learning-based moral value alignment problem.

*Non-learning-based Moral Value Alignment*: This approach involves analyzing the responses of PLMs without the need for learning or adaptation. The predefined function $M_\Theta$ is combined with a module that maps the responses of the PLMs to moral foundations based on the principles of MFT. This approach entails analyzing PLM responses without the need for learning or adaptation. The predefined function $M_\Theta$ is combined with a module that maps the responses of the PLMs to moral foundations based on the principles of MFT. For instance, PLMs can be treated as questionnaire respondents and their response can be compared with human responses, hence providing insight into the personality of PLMs.

## 5. Survey, Datasets, and Lexicon with Moral Values

The MFT is a cornerstone of datasets focusing on the morality field (Guo et al., 2023b). Insights from current studies on morality have primarily been obtained via analyzing moral vignettes, questionnaire data, and social media datasets (Hofmann et al., 2014). Therefore, we categorize the data sources into 3 categories: (i) **Moral surveys**, which correspond to set of close-ended questions designed to gauge individuals' moral beliefs, (ii) **Moral lexicons**, which refer to lists of word-score pairs related to moral values, and (iii) **MFT-annotated datasets**, which refer to data manually or (semi-) automatically labeled with the dimensions of the MFT. Figure 6 shows examples of these three types of datasets. Table 1 provides a summary of the data sources described in this survey, along with their category (moral surveys, moral lexicons, mft-annotated datasets) and their size in terms of rows. These datasets represent some of the most suitable resources existing at the time of our study for integrating PLMs and MFT.

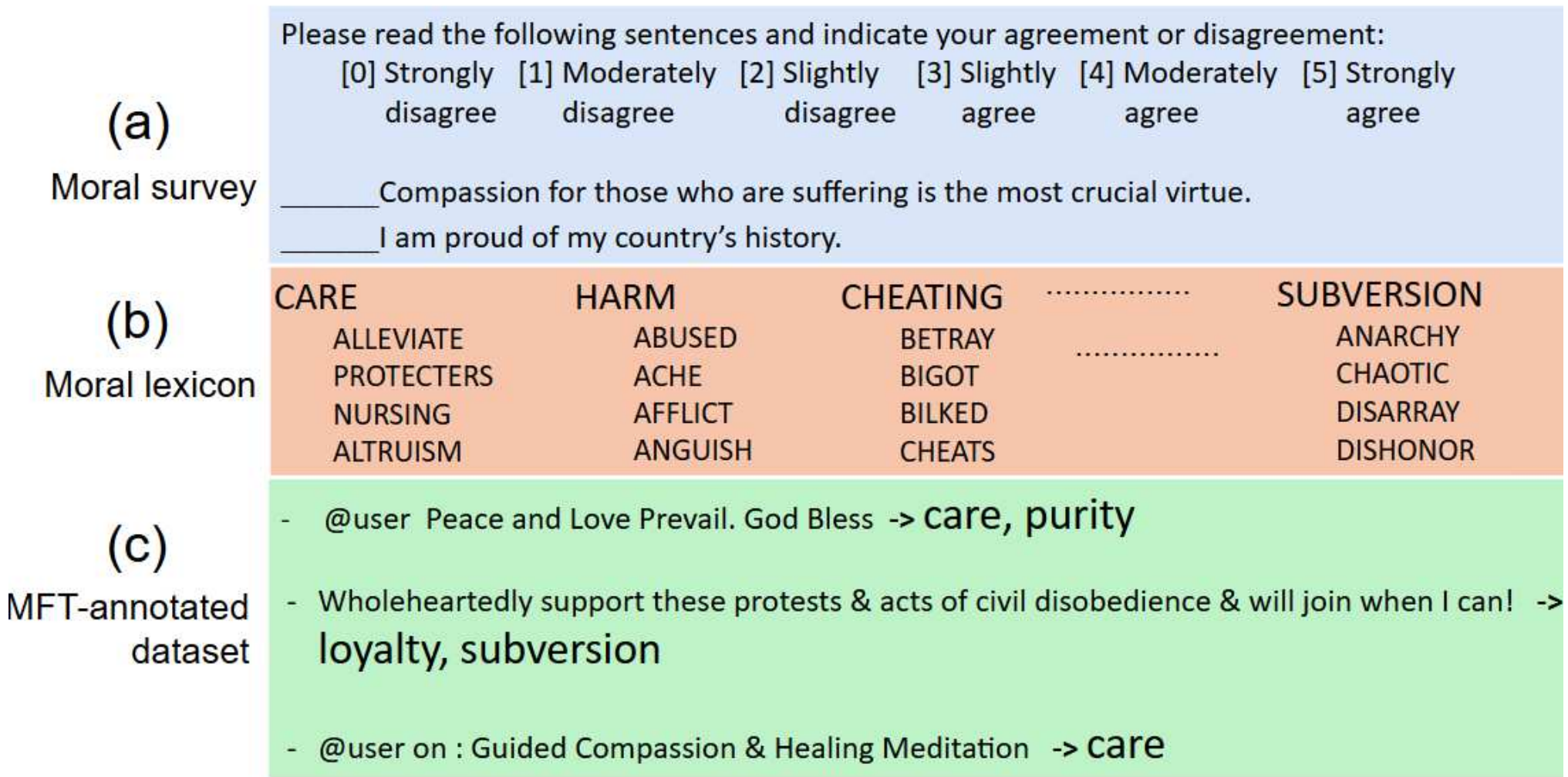

Figure 6: The types of datasets considered by the methods described in this survey: Moral surveys, moral lexicons and MFT-annotated datasets. (a) Example of Moral Survey dataset (extracted from the MFQ Graham et al. (2008)); (b) Example of Moral lexicon dataset (extracted from the MFDV2 Frimer (2019)); (c) Example of MFT-annotated dataset (extracted from the MFTC Hoover et al. (2020).

### *5.1. Moral surveys*

The Moral Foundations Questionnaire (MFQ) (Graham et al., 2008) is a psychometric tool designed to evaluate the extent to which individuals endorse the five different moral foundations of the MFT. It is widely used in social psychology to understand moral diversity across cultures and to examine individual differences in moral scenarios. The MFQ evaluates each foundation by utilizing 30 questions and six items, which are graded on a 6-point Likert scale (Likert, 1932), anchored from 1 (strongly disagree) to 6 (strongly agree). Table 2 shows moral foundations with corresponding example items statements used in the MFQ to assess an individual's moral values related to each foundation (Graham et al., 2008).

Table 1: Summary of the data sources used in the revised works, along with their category (Moral survey, Moral lexicon, MFT-annotated dataset) and their size.

| Dataset | Category | Size |
|---|---|---|
| MFQ (Graham et al., 2008) | Moral survey | 30 |
| eMFQ (Atari et al., 2023) | Moral survey | 36 |
| MFDv1 (Graham et al., 2009) | Moral lexicon | 324 |
| MFDv2 (Frimer, 2019) | Moral lexicon | 2014 |
| eMFD (Hopp et al., 2021) | Moral lexicon | 35, 985 |
| MoralStrength (Araque et al., 2020) | Moral lexicon | 1000 |
| MFTC (Hoover et al., 2020) | MFT-annotated dataset | 35,108 |
| MFRC (Trager et al., 2022) | MFT-annotated dataset | 16,123 |
| Covid (Rojecki et al., 2021) | MFT-annotated dataset | 2,648 |
| Congress (Johnson and Goldwasser, 2018) | MFT-annotated dataset | 2,050 |
| Extended Congress(Roy and Goldwasser, 2021) | MFT-annotated dataset | 161,295 |
| Facebook (Beiró et al., 2023) | MFT-annotated dataset | 4,498 |
| Facebook-2 (Kennedy et al., 2021) | MFT-annotatd dataset | 2,691 |
| Covid-19 (Pacheco et al., 2022) | MFT-annotated dataset | 750 |
| Exteded ArgQuality (Kobbe et al., 2020) | MFT-annotated dataset | 320 |
| MoralArg (Alshomary et al., 2022) | MFT-annotated dataset | 230,000 |
| MIC (Ziems et al., 2022) | MFT-annotated dataset | 38,000 |
| MAAFS (McCurrie et al., 2018) | MFT-annotated dataset | 69 |
| MFVs (Clifford et al., 2015) | MFT-annotated dataset | 132 |

Table 2: Example of items for each moral foundation in the Moral Foundations Questionnaire (MFQ) (Graham et al., 2008).

| Foundation | Example item |
|---|---|
| Care | "Compassion for those who are suffering is the most crucial virtue" |
| Fairness | "I think it's morally wrong that rich children inherit a lot of money while poor children inherit nothing" |
| Authority | "Respect for authority is something all children need to learn" |
| Loyalty | "People should be loyal to their family members, even when they have done something wrong" |
| Purity | "I would call some acts wrong on the grounds that they are unnatural" |
| Liberty | "The government interferes far too much in our everyday lives" |

The Extended Moral Foundations Questionnaire (EMFQ) (Atari et al., 2023) contains 36 questions built on top of the MFQ by incorporating the additional foundations of Liberty/Oppression, which address contemporary concerns about individual freedoms and economic rights, respectively. Like the MFQ, the EMFQ uses Likert scale assessments to gauge moral priorities. EMFQ has shown to be particularly useful for research in political psychology, allowing for a more descriptive analysis of ideological divides and moral conflicts in modern societies.

### 5.2. Moral Lexicons

The Moral Foundations Dictionary (MFD) (Graham et al., 2009) is the first vocabulary developed to assess moral values from textual data. The MFD provides a list of words and phrases associated with the five moral foundation dimensions. It has been used in various studies to understand cultural differences in moral values, political ideologies, and the language used in moral discourse (Kobbe et al., 2020).

The average number of words included in MFD for each moral category is 32, hence having a size of 324 words. MFDv2 (Frimer, 2019) is an extension of MDS, which consists of a lexicon of 2014 words, with each word associated to a single moral category. Another extension of the MFD is the extended Moral Foundation Dictionary (eMFD) (Hopp et al., 2021), a dictionary of 35,985 text samples extracted from English news articles on a variety of topics. Lemmas and stems are annotated with a moral value and words in the eMFD are weighted according to a probabilistic scoring procedure.

MoralStrength (Araque et al., 2020) is a semi-automated lexicon of approximately 1,000 lemmas, obtained as an extension of the MFD, which quantifies the relevance and strength of words related to the five moral foundations of the MFT. The lexicon is constructed by initially selecting seed words related to each moral foundation and then expanding this set through semantic similarity measures derived from word embeddings.

LibertyMFD (Araque et al., 2022) is in turn an extension of MoralStrength for detecting the Liberty moral dimension. Roy and Goldwasser (2021) extend the dataset proposed by Johnson and Goldwasser (2018) to build a topic indicator lexicon, for topics like Gun Control and Immigration.

### 5.3. MFT-annotated datasets

The Moral Foundations Twitter Corpus by Hoover et al. (2020) comprises 35,108 tweets annotated with 11 moral values based on the MFT. Each tweet is categorized into 10 moral values, reflecting the dyadic dimensions such as care/harm, which are treated separately, i.e., care and harm are separate labels. Additionally, the dataset includes a *non-moral* label for tweets that do not exhibit any moral sentiment, resulting in 11 possible labels per tweet. It also includes seven different domains related to morally relevant issues: (i) All lives matter, referring to American social movement that emerged as a reaction to the Black Lives Matter movement of the African-American community; (ii) The Baltimore Protests, concerning tweets related to the protest over the murder of Freddie Gray in Baltimore; (iii) Black Lives Matter, including all tweets with the hashtag #BlackLivesMatter; (iv) 2016 Presidential Election, consisting of tweets collected from followers of US politicians and reputable newspaper agencies throughout the 2016 presidential US election; (v) Davidson, a collection of hate speech and inflammatory language (Davidson et al., 2017); (vi) Hurricane Sandy, encompassing all tweets that were shared prior to, during, and shortly following the occurrence of Hurricane Sandy; (vii) METOO, which contains information from 200 persons who have participated in the social movement against sexual assault and harassment.

Further exploring the domain of social media, the Moral Foundations Reddit Corpus (MFRC) (Trager et al., 2022) introduces a dataset of 16,123 English Reddit posts drawn from 12 different subreddits, chosen according to the buckets of US politics, French politics and everyday moral life. Unlike Hoover et al. (2020), the MFRC does not deal with the Fairness/Cheating dimension of the MFT, but addresses the dimensions of Equality/Inequality and Proportionality/Disproportionality, following the framework of Atari et al. (2023). Equality/Inequality is a concept that promotes social justice and equality, while Proportionality/Disproportionality focuses on remuneration based on merit, contributing to virtues like productivity and deservingness, but also fostering corruption and nepotism. It also includes Thin morality and Non-moral categories, i.e., judgments which do not clearly refer to the six moral dimension, and judgments without moral content, respectively.

Beiró et al. (2023) gathered English Facebook comments focused on the topic of COVID-19 vaccination. They manually annotated 4,498 comments utilizing the same dimensions used by Hoover et al. (2020). In a similar fashion, Kennedy et al. (2021) also collected data from Facebook volunteers for analyzing individual moral concerns. Specifically, status updates of English-speaking participants were collected, with a total of 107,798 status updates collected from 2,691 individuals, along with their responses on the 30-item MFQ, consisting of two 15-item sections: the first measures the significance that individuals attribute to each of the five moral foundations, while the second contains contextualized elements designed to assess real moral judgments.

Johnson and Goldwasser (2018) collected tweets of US politicians related to 6 topics: (i) abortion, (ii) the Affordable Care Act, (iii) guns, (iv) immigration, (v) LGBTQ rights, and (vi) terrorism. They annotated 2,050 tweets based on topics, policy frames, and the five moral foundational dimensions of the MFT, plus a non-moral category, enabling to frame the problem as an 11-label classification task. Roy and Goldwasser (2021) further extended this dataset by collecting more tweets from US congress members, using a lexicon-based approach.

Pacheco et al. (2022) built a corpus of 750 tweets about the COVID-19 vaccine manually annotated by morality foundation, and vaccination stance (pro, anti, neutral, disagree). Similarly, Rojecki et al. (2021) annotated 2,648 tweets related to the COVID-19 pandemic.

Kobbe et al. (2020) augment the ArgQuality Corpus (Wachsmuth et al., 2017), which consists of textual debate arguments for two stances on 16 issues, such as evolution vs. creation and ban plastic water bottles. The dataset contains 320 arguments, covering 16 topics balanced for stance. Kobbe et al. (2020) annotates this dataset with the moral dimensions (grouped by dichotomy). The dataset can be leveraged to explore the interplay between moral values in arguments and various aspects such as argument quality, stance, and audience reactions. A related dataset is the one by Alshomary et al. (2022), which utilizes the lexicon provided by Hulpus et al. (2020) which connects moral foundations to Wikipedia concepts. This resulted in a balanced dataset across the five moral dimensions, with 230k argumentative texts annotated with moral values.

The Moral Integrity Corpus (MIC) (Ziems et al., 2022) was built up from Social Chemistry 101 (Forbes et al., 2020), which is a corpus of norms describing binary judgments of everyday situations. MIC extends Social Chemistry 101 by providing moral annotations related to the MFT on 38,000 prompt-reply pairs.

The Moral and Affective Film Set (MAAFS) (McCurrie et al., 2018) contains a collection of video clips based on the MFT dimensions, obtained by collecting and validating 69 videos from YouTube that represent various moral foundations. Such dimensions were rated by independent participants using standardized questionnaires which include: moral foundation (moral value from MFT), arousal (intensity of the emotional response), wrongness (degree to which the actions depicted in the videos are judged as morally wrong), weirdness (how strange or uncommon the moral acts depicted in the videos are perceived), clarity (how clear and understandable the moral content and context of the videos are) and so on.

The Moral Foundations Vignettes (MFVs) (Clifford et al., 2015) introduces a collection of moral vignettes, each violating one of the five moral foundations. It contains 132 vignettes, crafted to encompass diverse content and validated through factor analysis and human participants.

## 6. Methods

We organize existing studies into two main categories. The first one regards **Moral-driven PLMs**, which refer to works that adopt training strategies for learning model that predict or are aligned with moral values. The general approach here is to fine-tune or further pre-train PLMs for predicting the moral values of the underlying moral foundations hidden in the data. The second category, namely **Moral-targeted PLMs**, regards methods that seek to assess the moral values embedded into the PLMs without any moral task-specific adaptation, to probe their responsiveness to morally challenging issues and situations. The general approach is to analyze the PLM responses to determines their beliefs and the effects of training on a particular dataset. Figure 7 shows how these two categories are intertwined with various aspects related to the field of PLMs within the context of MFT. These aspects include the type of MFT-related tasks addressed (cf. section 4), the type of data employed (cf. section 5), the architecture of the models and the strategies employed for the resolution of the problem (cf. section 3). Solid lines indicate hierarchical (i.e. parent-child) dependencies within the tree diagram, while dashed lines indicate dependencies relationships. By examining the interrelationships between these factors within each category, we can identify specific correlations and patterns that allow us to organize and comprehensively describe the field of intersection between PLMs and MFT.

Table 3 shows the selected works for this survey. The table is partitioned into two subtables based on the aforementioned categories. For each method, we describe the utilized model(s) (Model column, cf. section 3.2); the addressed task (MFT-related task column, cf. section 4); the strategies for tailoring PLMs to the downstream application (Strategy column, cf. section 3.2.4); the data used for training and/or evaluation (Data column, cf. section 5); and the source code availability (Code column).

We propose a meta-definition for the problem of integrating PLMs with the MFT, aiming to establish a formal foundation framework for all the works examined. The proposed meta-definition must be broad enough to include both methods that use PLMs to learn hidden patterns related to the MFT (Moral-driven PLMs) and methods that use the MFT to analyze hidden patterns related to the PLMs' responses (Moral-targeted PLMs).

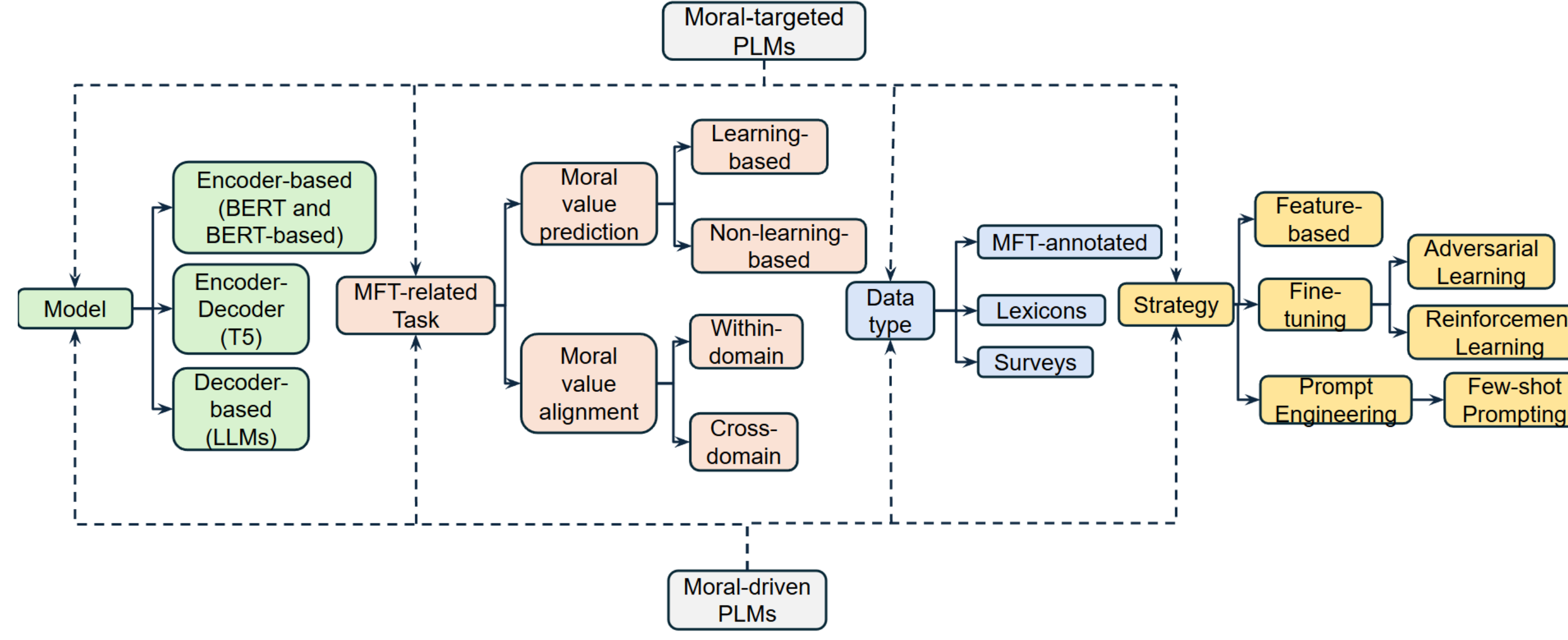

Figure 7: The interplay between the two categories of methods (Moral-targeted PLMs and Moral-driven PLMs) and various aspects related to PLMs within the context of MFT, which encompass: the architecture of the models (green rectangles), the types of MFT-related tasks addressed (red rectangles), the data employed (blue rectangles) and the strategies used for problem resolution (yellow rectangles). Dashed lines indicate dependency relationships, while solid lines indicate parent-child dependencies within the tree.

Table 3: Summary of the selected works in this survey

| Ref. | Model | MFT-related task | Data | Strategy | Code |
|---|---|---|---|---|---|
| **Moral-driven PLMs** | | | | | |
| Roy and Goldwasser (2021) | DRaiL + BERT | Within-domain moral values prediction | Twitter | Fine-tuning | |
| Pacheco et al. (2022) | DRaiL + BERT | Within-domain moral values prediction | Twitter | Fine-tuning | ✓ |
| Bulla et al. (2022) | SqueezeBERT | Within-domain moral values prediction | Twitter (MFTC) | Fine-tuning | |
| Rao et al. (2023) | BERT | Within-domain moral values prediction | Twitter (MFTC), Covid, Congress | Fine-tuning | |
| Trager et al. (2022) | BERT | Within-domain moral values prediction | Reddit (MFRC), Twitter (MFTC) | Fine-tuning | ✓ |
| Liscio et al. (2022) | BERT | Cross-domain moral values prediction | Twitter (MFC) | Further pre-training + Fine-Tuning | ✓ |
| Guo et al. (2023a) | BERT + neural architectures | Cross-domain moral values prediction | Twitter (MFTC, Covid, Congress) and news (eMFD) | Fine-tuning + Adversarial learning | ✓ |
| Preniqi et al. (2024) | BERT | Within-domain and cross-domain moral values prediction | Twitter (MFTC), Facebook, Reddit (MFRC) | Fine-tuning + Adversarial learning | ✓ |
| Kobbe et al. (2020) | S-BERT, BERT | Within-domain and cross-domain moral values prediction | MFD, Twitter (MFTC), Extended ArgQuality | Fine-tuning | ✓ |
| Alshomary et al. (2022) | BERT + Project Debater | Within-domain moral value prediction and arguments generation | MoralArg | Fine-tuning | ✓ |
| Dognin et al. (2024) | Open-Assistant 12 B | Learning-based moral values alignment | MIC | Fine-tuning + Reinforcement Learning | |
| **Moral-targeted PLMs** | | | | | |
| Abdulhai et al. (2023) | GPT-3 | Non-learning-based moral value alignment | MFQ | 1-shot prompting | ✓ |
| Simmons (2023) | GPT-3, GPT3.5 and OPT | Non-learning-based moral value alignment | Social scenario datasets + moral dictionaries MFDv1, MFDv2, and eMFD | 0-shot prompting | |
| Jiang et al. (2024) | DAMF + Social-PLM | Moral values prediction | Covid data | Feature-based | |
| Carrasco-Farre (2024) | Claude 3 Opus | Non-learning-based moral value alignment | persuasion dataset+ eMFD | 0-shot prompting | |
| Fraser et al. (2022) | T5 | Non-learning-based moral values alignment | MFQ | 0-shot prompting | ✓ |
| Kennedy et al. (2021) | BERT | Non-learning-based moral values alignment | Facebook-2 + moral dictionaries MFDv1, MFDv2 | Feature-based, | |
| Hämmerl et al. (2023) | S-BERT | Non-learning-based moral values alignment | MFQ | Feature-based | ✓ |
| Xie et al. (2020) | S-BERT | Moral value prediction | MAAFS, MFVs | Feature-based | |

*Meta-Definition (Integration of PLMs and MFT).* Let $M_\Theta$ denote a PLM with parameters $\Theta$, and $T_D$ denote the MFT equipped with a selected set of moral dimensions $D$ (e.g., the set of original moral dimensions proposed by Haidt and Joseph (2004), or alternative set). An integration process between a PLM $M_\Theta$ and a MFT $T_D$ can be described as any function belonging to a class of functions $\mathcal{F} = \{f \mid f : A \rightarrow B \times \mathbb{R} \ s.t. \ A, B \subseteq \mathcal{T} \setminus \{\emptyset\}\}$, where $\mathcal{T} = \{M_\Theta, \Psi_{T_D}\}$ and $\Psi_{T_D}$ is the set of possible non-empty outcome sets dependent on $T_D$, e.g., $2^D \setminus \{\emptyset\}$.

The above meta-definition serves as a basis for the specialized definitions used in this work: Moral-driven PLMs and Moral-targeted PLMs. The MFT's role is the key difference between the two categories: in moral-driven PLMs, the MFT informs the training process, enabling the model to predict moral values directly from input data without needing the MFT during inference; by contrast, moral-targeted PLMs utilize the MFT at inference time to prompt the PLM and to interpret the moral content of the model's responses. In the following, we specialize the integration function for either categories.

*Moral-driven PLMs.* PLMs are here adapted during the training stage to make inference based on the MFT. The integration process involves a PLM $M_\Theta$ in which knowledge about the MFT is embedded into the parameters $\Theta$. Moral-driven PLM can hence be expressed by a function $f \in \mathcal{F}$ such that:

$$f : M_\Theta \mapsto \Psi_{T_D} \times \mathbb{R}.$$

The process for learning the parameters $\Theta$ typically employs fine-tuning or pre-training techniques on data labeled with moral dimensions $D$. Note that when the parameters $\Theta$ are trained, as in the case of moral-driven PLMs, we can deal with the moral values prediction task or with the learning-based moral value alignment tasks. For example, in the former case, we have $f : M_\Theta \mapsto 2^D \times \mathbb{R}$, in which the integration of $\mathbb{R}$ takes into account the case in which we are dealing with both classification and regression tasks (e.g., ranking moral values).

As an example, consider a BERT model trained on a corpus of social media posts labeled with the Care and Harm moral values. When applied to new social media posts, the model can predict whether a post reflects a Care or Harm moral value, or to score how much a post contains Care and Harm moral values.

*Moral-targeted PLMs.* MFT is here used to analyze and assess the moral values embedded within the responses generated by PLMs. The integration process involves evaluating the responses of $M_\Theta$ in order to assess its behavior w.r.t. the MFT $T_D$. Moral-targeted PLMS can hence be expressed by a function $f \in \mathcal{F}$ such that:

$$f : M_\Theta \times \Psi_{T_D} \mapsto \Psi_{T_D} \times \mathbb{R}.$$

Note that in this case, the knowledge about the MFT is not embedded into the parameters $\Theta$, but it is injected through a transformation $\Psi_{T_D}$. The function $f$ takes the MFT $T_D$ as input and as a basis to analyze the output produced by $M_\Theta$, hence producing a response belonging to the subspace of $\Psi_{T_D} \times \mathbb{R}$. For instance, considering the Non-learning based moral value alignment task, an element of $\Psi_{T_D}$ in the domain corresponds to prompts with elements of $T_D$ (e.g., moral survey datasets). $\Psi_{T_D}$ in the co-domain can be the power set of $D$.

As an example, consider prompting GPT-3 with questions related to moral dilemmas, such as "Is it acceptable to lie to protect someone's feelings?" The responses generated by GPT-3 are then analyzed using the MFT to determine which moral dimensions, such as Fairness or Loyalty, are reflected in the answers.

### 6.1. Moral-driven PLMs

In this section, we describe moral-driven PLMs. Initially, we discuss methods that leverage BERT-based model for moral values prediction on social data in both within-domain and cross-domain scenarios. Then, we analyze studies that identify moral values to support argument analysis and moral values alignment.

*PLMs for moral values prediction in social discourses.* Bulla et al. (2022) extend the work of Hoover et al. (2020) by replacing the LSTM model with a PLM. Specifically, SqueezeBERT is fine-tuned on the labeled tweets of the MFTC dataset for the task of multi-label classification of moral values. Each of the moral dimensions has been defined as the union of the two moral values that compose it, e.g., for the Care moral value, positive tweets are the ones labeled either with Care or Harm. Experimental results show that the BERT-based model outperforms the LSTM in all seven domains of the MFTC, achieving an F1 score ranging from 47.0 to 86.0 across the moral dimension. Similarly, Rao et al. (2023) use a within-domain training approach to fine-tune BERT on Covid-19, MFTC and Congress datasets. The fine-tuned model is used to study partisan differences in moral languages related to Covid-19. One of the main findings is that conservatives use more negatively-valenced moral language than liberals, which indicates a more critical and morally charged tone in conservative tweets regarding Covid-19 issues. Also, political elites, regardless of their ideology, use moral rhetoric more extensively than non-elites, which suggests that leaders and influencers are more likely to frame Covid-19 issues in moral terms, potentially to mobilize their base or shape public opinion. Moreover, certain issues like masking and lockdowns may be more morally charged and salient in conservative discussions, while others like vaccines might be more prominent in liberal discourse.

Trager et al. (2022) fine-tune BERT on the moral dimensions of the MFRC, both in multi-label and binary classification settings. In the former case, all moral categories are predicted simultaneously. In the latter case, each label is predicted separately. Experiments have shown that multi-label BERT achieves lower performance than BERT as a binary classifier, with F1 scores ranging from 0.37 to 0.51. On the MFTC, achieved F1 scores range from 0.54 to 0.82. A cross-domain strategy is also considered, by down-sampling the MFTC dataset to have the same number of samples for each label as MRFC for consistency purposes. Then, the vanilla BERT model is trained on the MFRC and tested on the down-sampled MFTC and vice versa. Interestingly, models trained on the MFRC and tested on the MFTC have better classification performance, with F1 scores ranging from 0.28 to 0.53 vs. 0.31-0.43 across the moral dimensions. This is mainly ascribed to the Fairness Foundation's split into proportionality and equality, as described in Section 5.3, which makes cross-domain training more difficult. The effect of cross-domain classification on the moral prediction task have been thoroughly investigated by Liscio et al. (2022). Each dataset of the MFTC is treated as a different domain, thus obtaining seven datasets each corresponding to a topic (e.g., Black Lives Matter, Sandy, and so on), and train a BERT model on different settings to evaluate its generalizability and transferability capabilities for moral values prediction as well as its catastrophic forgetting property. Each setting varies based on the source of training data, evaluation, or fine-tuned data. Four different scenarios for multi-label classification are experimented, whereby the MFTC dataset is partitioned into source and target data. The former are six of the seven MFTC datasets treated as available data, while the latter is incoming dataset from a novel domain, always composed of one MFTC dataset. The settings are as follows: (i) in the *source* scenario, the source data correspond to the training set; (ii) in the *target* scenario, the training set corresponds to the target data; (iii) in the *fine-tune* scenario, the classifier is trained on the source data, and then fine-tuned on the target data; (iv) in the *all* scenario, the training set includes both source and target data. In each scenario, the classifier is evaluated on both source and target data, resulting in eight total combinations. Results reveal that the *all* scenario is in general the best one, obtaining 0.75 of F1 score when the target dataset is related to the topic 'ME TOO'. Moreover, similarly to Trager et al. (2022), pre-training a BERT classifier on the MFTC yields better performance on unseen moral domains, even when little training data is available. However, fine-tuning on a novel domain causes catastrophic forgetting of the domain it was pre-trained with, even when fine-tuning on a small portion of data from the novel domain.

Guo et al. (2023a) propose *DAMF* (Domain Adapting Moral Foundation inference model), a framework for moral value prediction on heterogeneous datasets, leveraging an adversarial training strategy. Rather than training the classifier on the aggregation of data, features from different datasets are mapped into a shared common embedding space. Specifically, a pre-trained BERT model is used to encode text, and a domain classifier is learned with the purpose of detecting the correct domain the data come from. The domain classifier works as an adversary, hence pushing BERT embeddings to be domain invariant (since they have to "fool" the domain classifier). Then, moral foundations are classified based on the learned domain-invariant embedding. DAMF is evaluated using different datasets in the cross-domain setting: Covid-19, Congress, MFTC and the eMFD. Results show that model performance improves when fusing heterogeneous datasets. When using Covid-19, Congress and eMFD as training data, the achieved F1 score performance on the MFTC is equal to 0.43.

MoralBERT (Preniqi et al., 2024) is a framework in which both a base BERT and a BERT trained with the adversarial learning strategy used by Guo et al. (2023a) are fine-tuned for single-label and multi-label moral values prediction. Similarly to Trager et al. (2022) and Liscio et al. (2022), the former setting outperforms the latter in terms of F1 score. MoralBERT with adversarial learning outperforms MoralStrength by a large margin and achieves F1 score equal to 0.65 and 0.52, in the single-label and multi-label cases, respectively. The performance degradation in the multi-label scenario likely due to the fact that BERT model is fine-tuned on a specific moral value (e.g., Care). In a cross-domain scenario (training on two social media datasets, and testing on the third left-out social media dataset), the overall performance tends to decrease since the distribution of moral values differs from dataset to dataset, even if they are from the same domain.

Methods such as Roy and Goldwasser (2021) and Pacheco et al. (2022) integrate PLM encoders within declarative frameworks, by associating a BERT classifier to logical rules. Both works exploit DRaiL (Pacheco and Goldwasser, 2021), a framework that combines logical reasoning and deep learning for addressing NLP tasks. Roy and Goldwasser (2021) aim to use MFT for elucidating the political stances of American politicians and analyzing their positions. This is addressed by using logical rules within DRaIL to model tweets and other meta-data, from the Extended Congress dataset. Each rule corresponds to a classifier that maps features on the left-hand side of the rule to the predicted output on the right-hand side. Three rules are defined to describe the foundation of a tweet, the ideology, and the topic, e.g. a tweet $x_i$, with a topic $t$, has a moral foundation $d_j$. Each base rule is associated with a BERT model to encode tweets. Combining multiple rules encoded with BERT is shown to be useful in prediction, achieving the highest performance in terms of the F1 score (0.52). Concerning the MFT, a strong correlation is found between the usage of MFT and politicians' nuanced stances on particular divisive issue, based on Twitter. For example, considering the topic of 'immigration', Democrats employ the term Care to refer to dreamers and young people, while Republicans use this term when discussing 'border wall' and 'border patrol'. Pacheco et al. (2022) utilize a similar architecture to identify the moral foundation in tweets about the Covid-19 vaccine. Each tweet is annotated with the moral foundation and the vaccination stance. Relational properties of tweets with logical rules are described, indicating whether a tweet has a moral judgment, its moral foundation, and its vaccination stance and, similarly to Roy and Goldwasser (2021), tweets are encoded using BERT. Jointly modeling moral judgment and vaccination stance is shown to lead to improvements in the F1 score performance on the moral prediction task.

*PLMs to support argument analysis.* The approaches by Kobbe et al. (2020) and Alshomary et al. (2022) aim to identify moral values for supporting argument analysis. Kobbe et al. (2020) propose different variants of S-BERT to generate text encoding. In one of these, the MFD is used as seed data, and S-BERT is fine-tuned to encode moral foundations in DBpedia[2] data labeled utilizing the MFD through a weak-supervision strategy. In another approach, S-BERT is enhanced with sense-disambiguation features through a word sense-disambiguated version of the MFD based on WordNet synsets. For evaluating the methods, both the MFTC and the extended ArgQuality corpus are used. In one setting, MFTC is used for training different models in a within-domain scenario. When dealing with a cross-domain scenario, models are trained on the MFTC and tested on the extended ArgQuality corpus. In both cases, multi-label BERT is the best-performing method, with an average F1 score equal to 0.45 and 0.308, respectively. The lower performance for the moral class would be explained by the differences in class distribution between the two datasets. Interestingly, in the experiments related to the correlations of multi-label BERT between arguments and moral values, a weak positive correlation is observed between argument quality and moral sentiment for the two most frequent categories on the ArgQuality corpus.

Alshomary et al. (2022) propose Moral Debater to examine how morally framed arguments influence different audiences. A model is learned to identify moral foundations in arguments by fine-tuning BERT on the MoralArg dataset. By testing it on the ArgQuality corpus, the model is found to outperform the multi-label BERT model of Kobbe et al. (2020), obtaining an average F1 score of 0.40 vs. 0.28. The learned model is used to study the effect of morally framed arguments on the audience by combining it with IBM's Project Debater (Slonim, 2022). Given as input a controversial topic, a stance on the topic, and a set of morals to be targeted, it generates arguments with the given stance. The findings indicate that the generated arguments are aligned with the audience's moral beliefs. In the case of conservative views, arguments that emphasize loyalty, authority, and purity are compelling, whereas for liberal views, arguments respond strongly to themes of care and fairness.

*Adapting PLMs to support specific moral values.* Considering the task of learning-based moral value alignment, Dognin et al. (2024) propose Contextual Moral Value Alignment Through Context-Based Aggregation (CMVA-GS), which leverages the MFT to create a series of moral agents based on OpenAssistant 12B. Each moral agent is fine-tuned using reinforcement learn-ing, with reward functions tailored to its respective moral values. The reward function indicates the reward for a specific action at a given prompt based on the moral value being optimized. The moral agents' responses are then aggregated contextually to generate the final output. The fine-tuning and evaluation dataset is the Moral Integrity Corpus (MIC), which is labeled on five moral dimensions. The CMVA-GS models tend to achieve the highest ROUGE scores, indicating better alignment with moral values compared to other PLMs such as Llama-7B and Llama-13B.

[2]`https://www.dbpedia.org/`

### *6.2. Moral-targeted PLMs*

We further explore the existing literature of works examining the moral alignment of PLMs. We firstly review the literature that leverages LLMs, which inherit moral inclinations during their training phase or are influenced by prompt design to provide biased responses in relation to specific moral values. Then, we delve into studies which explore the inherent moral biases in the contextual embeddings of encoder-decoder and encoder-based PLMs.

*Moral values discovery through prompt engineering.* Abdulhai et al. (2023) conducted a series of experiments analyzing the moral foundations of GPT-3 models (*curie*, *babbage* and *davinci*) to understand the values encoded from their training data and their potential impact on unforeseen tasks. The models are examined to understand whether they can exhibit specific moral stances, the consistency of these stances across various contexts, and the ability to deliberately prompt them to endorse particular moral foundations. Questions from the MFQ are exploited to construct prompts containing a description of the task, a rating scale specifying the model response (ranging from 0 to 5, with 0 corresponding to 'not relevant' and 5 to 'extremely relevant'), and a static example to show the model how to structure its response. To evaluate potential cultural and political biases in GPT-3, its default responses are analyzed and prompted with a political affiliation. By comparing the response of GPT-3 models with human studies, it is observed that less expensive GPT-3 models (*curie*, *babbage*) exhibit greater differences from human moral foundations, while more expensive models (*davinci)* align more closely, suggesting that larger models may better capture moral political values. As regards the consistency of PLMs' moral foundations across different contexts, certain dimensions (e.g., Fairness and Purity) are weighted more strongly than others. Additionally, the impact of prompting the models on exhibiting a particular moral stance is studied: results reveal moral correlations in the model's responses (prompts exalting Authority also lead to higher scores on the Purity traits, similarly for Fairness and Harm). Overall, the results suggest that while GPT-3 models may exhibit certain political tendencies, these can be modified through deliberate prompting, which also affects behavior in downstream tasks; for example, models prompted to prioritize different moral foundations exhibit significant behavioral differences in donation decisions.

Simmons (2023) examines the "moral mimicry" phenomena, i.e., whether PLMs can replicate the US political party moral biases. The goals are (i) to assess whether PLMs utilize moral terminology appropriately in different situations; (ii) if they can emulate human moral biases when given a political identity; and (iii) if and how the model size influences such capability. Prompts are intended to motivate language models to provide clear moral justifications. Each prompt contains a scenario, a political identity statement and a moral stance. The scenario is a description of circumstances or behaviors that need moral assessment, extracted from three datasets (Moral Stories (Emelin et al., 2021), ETHICS (Hendrycks et al., 2021), and Social Chemistry (Forbes et al., 2020)). A political identity statement is a text referencing political beliefs, such as "liberal" or "conservative". A moral stance labels the scenario's acceptance or disapproval such as "moral" or "immoral". Five distinct prompt templates are utilized to assess sensitivity to specific phrasings. Three dictionaries (MFDv1, MFDv2, and eMFD) are used to evaluate the moral foundational content of the responses from GPT-3, GPT-3.5, and OPT models. Specifically, these dictionaries are applied to the responses to generate scores indicating the presence of each moral foundation. To focus on foundational content, the positive or negative connotations of words are removed, merging Virtue and Vice words into a single category (for example, a category is Care/Harm). Also, varying the political identity in prompts changes the likelihood of expressing a particular moral foundation. Experimental results regarding the first goal indicate that the PLMs increase their use of a specific moral foundation when it is salient in a given scenario. Additionally, PLMs diverge more from human consensus compared to individual humans, suggesting that PLMs do not perfectly replicate human use of moral foundations. When conditioned with political identity, the results reveal that PLMs employ more individualizing foundations (Care/Harm and Fairness/Cheating) when prompted with a liberal identity and more binding foundations (Authority/Subversion,

Sanctity/Degradation, and Loyalty/Betrayal) when prompted with a conservative identity. Finally, larger models generally exhibit stronger moral mimicry capabilities, particularly within the OPT family. Jiang et al. (2024) utilize DAMF in combination with Social-PLM (Jiang and Ferrara, 2024) for detecting moral values on a Covid-19 dataset (Chen et al., 2020), and analyzing user groups in social media. Users who value fairness and authority engage more frequently in discussions, while those emphasizing Care and Purity show distinct participation patterns. Also, tweets combining multiple moral foundations are rare but receive more engagement.

Carrasco-Farre (2024) examines the ability of PLMs to adjust their arguments based on the prompt, employing persuasive strategies that require different levels of cognitive effort and utilizing different moral language. To this purpose, the study utilizes the responses of the experiments performed by Anthropic [3], in which human participants were presented with various claims and corresponding persuasive arguments generated by a PLM (specifically Claude 3 Opus) or human-generated. Persuasiveness is measured by shifts in agreement with claims before and after exposure to arguments. The dataset included 56 diverse claims on divisive topics such as police body cameras and lab-grown meat. The PLM was prompted to produce arguments using different persuasive techniques (for example, arguments aimed at persuading undecided or opposed individuals or arguments crafted with emotional appeal, logical reasoning, and credibility). The analysis focuses on both cognitive effort, through readability and perplexity, and the moral-emotional language, evaluated using sentiment analysis and the frequency of words expressing MFT moral traits. The PLM used more moral language than humans overall, with a higher frequency of both positive and negative moral foundations, contributing to its persuasive power. This aligns with previous research indicating that moral-emotional language captures attention and enhances persuasiveness. The study suggests that the strategic use of moral language by PLMs, including negative bias such as harm and cheating, plays a significant role in persuasion.

*Moral bias analysis in PLMs embeddings.* Fraser et al. (2022) explore the Delphi framework for the MFT. Delphi (Jiang et al., 2021) is a PLM built on the UNICORN (Lourie et al., 2021) model, a specialized version of T5 for commonsense question answering, and is fine-tuned on the COMMON-

[3] `https://www.anthropic.com/news/measuring-model-persuasiveness`

SENSE NORM BANK dataset (Jiang et al., 2021), containing 1.7 million ethical judgments. Specifically, it consists of various moral judgments on everyday situations, constructed by collecting and annotating a large number of diverse scenarios that might morally and socially salient situations. Delphi is trained on the COMMONSENSE NORM BANK to respond to ethical queries in various formats, predicting moral judgments that align with general societal views. The system's final task is to provide moral judgments on everyday situations in response to user queries, offering answers in yes/no, free-form, or relative modes based on the ethical implications of the described situations, showing a high level of accuracy w.r.t. GPT-3 in predicting correct moral judgement. According to Fraser et al. (2022), when using the MFQ as a basis for their Delphi queries, the two most crucial foundations are Care and Fairness. These are also called the individualizing foundations, in contrast to the other three foundations, which are called the binding foundation (Graham and Haidt, 2010). Loyalty, authority, and purity, the three pillars upon which our society rests, take a slightly lower position. Although both Authority and Loyalty are commonly linked to the Community ethic, the order of importance is reversed here, with Authority being placed higher than both of them. Authority, on the other hand, is associated with religious hierarchy and tradition, which can lead to its connection to the Divinity ethic.

Kennedy et al. (2021) aim to assess a direct connection between individual moral concerns and naturally observed language, examining both moral and general language (i.e., not necessarily expressing moral concerns). In this study, data are gathered from 2,691 participants, including their responses to the MFQ and 107,798 Facebook posts. Two main approaches are used to obtain a text representation expressing the relation between moral concerns and naturally observed language. The first is MFD-based, testing the presumed link between moral language through the MFD and MFDv2, and individual moral concerns. The relation between moral and natural language is measured calculating the rate of occurrence per moral category, i.e., the frequency of a word from the moral category w.r.t. the total number of words in the social posts of a given participant. The second approach is general language-base, consisting of bag-of-words modeling with LDA and text embedding with GloVE and BERT. A regression analysis is carried out to predict each MFQ score from the text representations. Experimental results reveal that MFD-based measures are generally less effective at predicting moral concerns compared to NLP-techniques. In particular, BERT was able to capture differences in language w.r.t. moral concerns that were not captured by individuals' lexicons.

Xie et al. (2020) introduce a text-based method for predicting human moral judgments of moral scenarios. The aim is to ascertain whether human-judged moral categories can be deduced solely from text, leveraging contextualized language models. The study assesses different representations of moral scenarios and their efficacy in categorizing nuanced moral differences, even with limited training data. The scenarios are gathered from three independent datasets, including MAAFS and MFVs. The experimental settings evaluate several classifiers (Gaussian Naive Bayes, Regression, KNN and SVM) with different representations of moral vignettes, including contextual embeddings with S-BERT. S-BERT embeddings achieve the highest accuracy across all datasets and classification methods (mean accuracy 0.44 for MAAFS and 0.57 for MFVs).

Hämmerl et al. (2023) investigate whether moral standards learnt from dominant language sources (e.g., English) are imposed on other languages during cross-lingual transfer learning, possibly leading to negative consequences. To this aim, the MORALDIRECTION framework for multilingual PLMs is utilized to encode action verbs with positive and negative connotations using a sentence embedding model. These verbs are placed in template questions to generate sentence embeddings, with PCA used to derive a "moral direction" subspace. Scores are normalized between [-1, 1], indicating moral value estimations. The framework is evaluated with both multilingual (e.g., mBERT, XLM-R) and monolingual models, followed by S-BERT models trained for each language. Results show high correlation between S-BERT models and human moral judgments across languages, though slightly lower for Arabic and Czech in the XLM-R model. The study finds that training with parallel data eliminates cultural differences, aligning models closely with English. Analysis of parallel sentence pairs reveals discrepancies due to literal translations and lexical confusion, especially in Czech-English pairs. Moreover, PLMs' moral dimensions are evaluated using the MFQ, by comparing the model scores against human responses from various studies to assess how well models capture human moral intuitions across different cultures. Models' responses are simplified for better comprehension, and their scores are mean-pooled within each moral foundation category. Findings reveal that models can generally discern right from wrong but show inconsistencies on sentences that are complex and with negation proposition. The models' scores do not align perfectly with human responses and vary across languages. This discrepancy is more pronounced when models are trained on smaller datasets for specific languages, indicating the influence of training data on moral judgments. However, German and Chinese languages show high agreement with English, and all languages correlate strongly in the pre-existing S-BERT model trained on parallel data, i.e., sentence pairs expressed in different languages, suggesting that training with parallel data promotes more uniform behavior, reducing cultural disparities.

## 7. Findings, challenges and future directions

Our discussion so far has provided insights into the potential of PLMs in modeling the framework of the MFT. These insights might contribute to a deeper understanding of the complexities involved in integrating MFT and PLMs and the ongoing efforts to improve their performance.

*Main trends in the existing literature.* Observing Table 3, recurring patterns can be noticed regarding models selection, tasks addressed, data employed, and strategies used (the interplay among the categories and these aspects are shown in Figure 7). From the perspective of model choice, BERT and BERT-based models represent the most commonly used technology when creating moral-driven PLMs, while LLMs are primarily employed as moral-targeted PLMs. Clearly, the choice of the model has a significant impact on the adaptation strategy. Therefore, a clear separation of strategies can be observed between moral-driven PLMs, which exclusively use fine-tuning strategies to adapt to the MFT task, and moral-targeted PLMs, which adopt few-shot prompting and feature extraction. Moral-driven PLMs are frequently employed for tasks involving moral values prediction, while moral-targeted PLMs are predominantly utilized for moral value alignment tasks. Specifically, MFT-annotated datasets are generally used in the context of within-domain and cross-domain moral values prediction tasks. Conversely, lexicons and moral surveys are primarily used for addressing moral value alignment challenges. As regards the employed data, we can notice that moral surveys and moral lexicons are mostly used to evaluate moral value alignment, while MFT-annotated datasets are primarily involved in tasks of moral values prediction.

*Correlation between moral foundations and political stances.* A prominent theme is the correlation between moral foundations and political stances. Roy and Goldwasser (2021) demonstrate a strong link between the use of moral foundation language in tweets and the nuanced political positions of U.S. politicians on issues such as gun control and immigration. This correlation is further explored by the work of Simmons (2023), who finds that changing the political identity in prompts affects the likelihood of expressing particular moral foundations, with PLMs using more individualizing foundations when prompted with a liberal identity, and more binding foundations when prompted with a conservative identity. Additionally, Rao et al. (2023) highlights partisan differences in moral language, noting that conservatives tend to use more negative moral language compared to liberals, suggesting a more critical and morally charged tone in conservative tweets.

*PLMs vs Lexicon-based methods.* The superiority of PLMs over traditional and lexicon-based methods is evident in several studies. Bulla et al. (2022) notice enhanced performance with PLMs upon earlier works, while Xie et al. (2020) highlight the superior performance of contextual embeddings in predicting human moral judgments. Additionally, Kennedy et al. (2021) point out the limitations of lexicon-based measures, suggesting that predefined moral dictionaries may not fully capture the complexity and variability of moral language as expressed naturally by individuals. The moral nuances in natural language can be better captured by PLMs. Furthermore, Roy and Goldwasser (2021) and Pacheco et al. (2022) investigated the integration of PLMs with logical reasoning frameworks. They show that combining PLMs such as BERT with declarative frameworks like DRaiL enhances the performance of NLP tasks by effectively modeling complex relational properties and dependencies in text data. This approach allows for the definition of logical rules that capture these dependencies, leading to improved performance.

*Effect of training strategies.* Training strategies, including fine-tuning and adversarial learning, have shown several enhancements in moral value prediction tasks. The works proposed by Kobbe et al. (2020), Alshomary et al. (2022), Dognin et al. (2024), and Guo et al. (2023a) highlight the effectiveness of these approaches in aligning model outputs with desired moral values. Fine-tuning emerges as the predominant strategy for adapting models to specific tasks and datasets, involving additional training on task-specific datasets to capture nuances and improve performance. Adversarial learning, as employed by DAMF (Guo et al., 2023a) and MoralBERT (Preniqi et al., 2024), is used to create domain-invariant embeddings, improving cross-domain performance by reducing the impact of domain-specific biases. Prompt engi-neering and reinforcement learning are notable strategies for aligning PLMs with specific moral values without extensive fine-tuning.

*Effect of prompting.* Several works have examined how the construction of prompts can elicit specific moral traits and influence PLM responses. This aspect is crucial for understanding how PLMs reflect and generate moral judgments, providing insights into the moral tendencies embedded in the models' pre-training data. Abdulhai et al. (2023) and Simmons (2023) show that deliberate prompting can significantly influence the moral foundations exhibited by models, affecting their responses and behaviors in various tasks. This highlights the malleability of PLMs and their potential to be guided through carefully designed prompts to exhibit desired moral and ethical considerations. Additionally, modifying the models' moral stances through prompting affects their behavior in downstream tasks, such as donation decisions, demonstrating the practical implications of moral prompting.

*Effect of cross-domain strategies.* Cross-domain strategies and their impact on model performance have been extensively explored. Trager et al. (2022), Liscio et al. (2022), Preniqi et al. (2024), Guo et al. (2023a) and Alshomary et al. (2022) demonstrate that training models on heterogeneous datasets generally improves performance. However, challenges remain due to class distribution differences and the phenomenon of catastrophic forgetting when models are fine-tuned on new domains, as noted by Liscio et al. (2022). This issue highlights the difficulty of maintaining model accuracy across varied datasets.

*Cross-lingual approach.* Cross-lingual transfer of moral standards is another critical area of investigation. Hämmerl et al. (2023) explores whether moral standards learned from dominant language sources (e.g., English) are imposed on other languages during cross-lingual transfer learning, which could lead to negative consequences. Training with parallel data tends to eliminate cultural differences, aligning models more closely with English, but models trained on smaller, language-specific datasets show more variability in moral judgments. Comprehensive training on diverse datasets is crucial for improving the alignment of PLMs with human moral values across different languages and cultures.

*The role of model size.* Model size also plays a crucial role in moral alignment. The works proposed by Abdulhai et al. (2023) and Simmons (2023) indicate that larger models, such as GPT-3, align more closely with human moral foundations compared to smaller models. This suggests that model complexity contributes to better capturing and representing moral values.

*Bias on moral judgement.* The potential bias in moral judgments is highlighted by Fraser et al. (2022), who find that moral judgments are influenced by the moral ideals of the annotators. This suggests that the system's responses may not be universally applicable or unbiased, highlighting the need for careful consideration of the demographic and ethical diversity of annotators in training datasets to ensure balanced and fair moral judgments.

*Moral labeling discrepancy and reliability of manually annotated data.* Another limitation regards the data sources utilized in the field (Table 1). Among the various datasets, a notable discrepancy is observed in the choice of labels, with some datasets employing 10 or more labels and others opting for 5. The underlying motivation for this decision is that both labeling schemes are closely related, and it is often ambiguous which end of the dimension is being addressed, particularly in the context of negated sentences; for instance, the sentence "I could never hurt you" could either be classified under Harm due to its use of vocabulary related to this dimension or be annotated as the opposite, Care, since it expresses an inability to harm someone, thereby aligning more strongly with the Virtue class (Kobbe et al., 2020). This duality underscores the complexity and subjectivity inherent in labeling and categorizing moral dataset, since also PLMs exhibit inconsistencies, especially with complex phrases and negation (Hämmerl et al., 2023).

A related issue is the reliability of manually annotated data. Before moral values can be reliably utilized for argument analysis, the quality of the annotations must be ensured. The low inter-annotator agreement for Moral Foundations annotation casts doubt on the validity of the findings, raising concerns about the consistency and accuracy of these annotations (Kobbe et al., 2020; Fraser et al., 2022). Therefore, there is an emergency for robust and standardized annotation processes to enhance the reliability of dataset annotations in moral value analysis.

*Lack of inclusive datasets.* Another related challenge is the construction of large-scale, multi-cultural, and multi-domain datasets. The majority of existing datasets are sourced from social media platforms, as indicated in section 5, which limits the diversity of language and context represented. The reliance on social media data, primarily in English, does not capture the full spectrum of moral values across different cultures and languages. To address this, it is imperative to develop multi-lingual datasets, especially incorporating low-resource languages which do not have the extensive data available for English, a high-resource language. Johnson et al. (2022) demonstrate the challenges associated with low-resource languages, where limited data availability hinders the training and performance of PLMs. Creating balanced and representative datasets from various cultural and linguistic backgrounds will ensure that PLMs can better generalize moral judgments across different contexts.

*Ontologies to guide PLMs.* Given the flexibility of PLMs and their significant responsiveness to prompting, there is a clear need for models that can be guided by predefined ontologies to provide consistent and coherent responses aligned with a specific worldview. The core issue here is that morality is deeply influenced by cultural factors, which vary widely across different societies. By integrating ontologies, which encapsulate structured knowledge and worldviews, models can be better directed to align their outputs with culturally relevant moral standards. This approach would involve embedding cultural and moral frameworks directly into the model's training process, ensuring that the generated responses adhere to these predefined moral principles. Such a method would mitigate the influence of arbitrary prompts and steer the model towards more stable and culturally coherent moral judgments.

*Explainability and interpretability.* A critical area lacking in current research is the explainability and interpretability of PLM outputs concerning moral judgments. Most existing works do not explore the reasoning behind the model's outputs, leaving a gap in understanding how decisions are made. This lack of transparency can undermine trust in these models, especially in sensitive applications involving moral and ethical considerations. Incorporating methods similar to the Chain-of-Thought approach, as demonstrated in the work of Jin et al. (2022), can significantly enhance the interpretability of PLMs. By explicitly detailing the steps and reasoning paths taken to arrive at a given solution, users can gain insights into the underlying processes of the model. This not only improves trust and reliability but also allows for the identification and correction of biases and errors in the model's reasoning. Developing frameworks for explainability and interpretability in moral prediction tasks is essential for advancing the field and ensuring the ethical deployment of PLMs.

## 8. Conclusion

This survey explored the field related to the intersection between MFT and PLMs, highlighting the efforts made in targeting and driving PLMs in the context of MFT. Furthermore, We formalized the MFT-related tasks addressed by current PLMs and provided a meta-definition that can encompass a broad spectrum of PLMs integrating the MFT. This theoretical framework enable to categorize the current approaches in the field.

Various models and strategies, including fine-tuning and prompt engineering, have shown promise in accurately predicting moral values and aligning PLMs within specific moral scenarios. However, challenges such as the reliability of annotated data and the performance of models in cross-domain scenarios remain.

The findings from the reviewed works reveal several key insights. Firstly, there is a notable correlation between moral foundations and political stances, demonstrating how PLMs can be influenced by and reflect societal biases. The superiority of PLMs over traditional and lexicon-based methods in capturing the complexity of moral language has been established, highlighting the potential of PLMs to better understand and generate morally nuanced content. Integration techniques, such as combining PLMs with logical reasoning frameworks and employing adversarial learning, have shown significant advancements in moral value prediction tasks. However, the need for comprehensive and standardized annotation processes is critical to enhance the reliability of moral datasets. Additionally, constructing large-scale, multicultural, and multi-domain datasets is crucial for improving the generalizability of PLMs in MFT-related tasks. Prompt engineering has been identified as a powerful tool for influencing PLM responses to align with desired moral values. This malleability underscores the importance of careful prompt design and the potential for integrating predefined ontologies to guide PLM outputs consistently. Furthermore, cross-lingual and cross-domain challenges highlight the need for diverse and representative training data. Ensuring that PLMs can generalize moral judgments across different cultures and languages is crucial for developing universally applicable morally aware AI systems.

In conclusion, the analysis presented in this survey provided a comprehensive overview of the current state of research, highlighting significant advances and ongoing challenges. By addressing these challenges and exploring new research directions, such as enhancing dataset reliability, improving cross-domain generalization, and integrating ontologies for consistent moral alignment, the field can move towards developing morally aware AI systems capable of operating effectively in diverse and multicultural environments. These advancements will pave the way for AI systems that are not only technically proficient but also ethically sound, contributing positively to the society.

## Acknowledgements

L.Z. and C.M.G. were visiting the University of Lausanne mostly during the development of this work.